%% file: transfer_in_smart_city.tex
\documentclass[12pt, journal, peerreview]{IEEEtran}
\usepackage{textcomp}
\usepackage{subfig}
\usepackage{amsthm}
\usepackage{amsmath}
\usepackage{multibib}
\usepackage[monochrome]{xcolor}
\usepackage{multirow}

\ifCLASSINFOpdf
  \usepackage[pdftex]{graphicx}
  \graphicspath{{./figures/}}
  \DeclareGraphicsExtensions{.pdf,.jpeg,.png}
\else
\fi

\begin{document}
\title{Smart City Development with Urban Transfer Learning}

\author{
	Leye Wang,
	Bin Guo,
	Qiang Yang
}

\maketitle

\setcounter{page}{1}

\section*{abstract}

Nowadays, the smart city development levels of different cities are still unbalanced. For a large number of cities which just started development, the governments will face a critical cold-start problem: `\textit{how to develop a new smart city service with limited data?}'. To address this problem, transfer learning can be leveraged to accelerate the smart city development, which we term the  \textit{urban transfer learning} paradigm. This article investigates the common process of urban transfer learning, aiming to provide city planners and relevant practitioners with guidelines on how to apply this novel learning paradigm. Our guidelines include common transfer strategies to take, general steps to follow, and case studies in public safety, transportation management, etc. We also summarize a few research opportunities and expect this article can attract more researchers to study urban transfer learning.

\textbf{keywords}: transfer learning, urban computing, smart city

\input{introduction}
\input{background}

\input{what_to_transfer}
\input{applications}
\input{challenges}

\bibliographystyle{unsrt}
\bibliography{transfer_in_smart_city}

\end{document}

%% file: introduction.tex
\section{Introduction}

Nowadays, smartphones, vehicles, and infrastructures (e.g., traffic cameras, air quality monitoring stations) are continuously generating a huge amount of urban data in heterogeneous formats such as GPS points, tweets, and road traffic. This opens up new opportunities to learn about city dynamics from a variety of perspectives and facilitates various smart city applications for traffic monitoring, public safety, urban planning, etc.

Meanwhile, smart city development levels of different cities are unbalanced. According to a progress report of China smart cities in 2014, although most first-tier cities have extensively employed smart city services, the percentages for the second- and third-tier cities employing smart city services  are just 65\% and 18\%, respectively \cite{liu2014china}.
For the governments of a large number of cities that just started smart city development, one key question emerges, `\textit{how to develop a new smart city service with limited data?}'. As an example, suppose a city plans to build a public safety early warning system based on crowd density to prevent potential crowd disasters such as stampede~\cite{zhang2017deep}, but there are few historical crowd flow records. Then, how can crowd flow be predicted for early warning without adequate data? To overcome data scarcity, current smart city practitioners usually have to first build a large-scale platform to collect and integrate much data before actually implementing specific smart city services. This means, before a city sees the benefits from state-of-the-art big data techniques, the governments and the related corporations must spend time and money for data collection. More seriously, this initial spending may be unguided as no one clearly knows which parts of data need to be prioritized. Can we  alleviate this difficulty and help bootstrap new smart city services more efficiently? 

In this article, we investigate the \textit{urban transfer learning} paradigm, a novel cross-discipline research area on applying transfer learning to address smart city cold-start problems. \textit{Transfer learning}~\cite{pan2010survey} is a series of machine learning techniques that transfer knowledge from a \textit{source domain} (with rich data) to a \textit{target domain} (with little data). Transfer learning has witnessed a lot of success stories in tasks such as text classification~\cite{pan2011domain} and product recommendation~\cite{li2009can}, but applying it to smart city is yet to be fully explored.
To the best of our knowledge, this is the first article to systematically study the \textit{urban transfer learning} paradigm with a focus on common issues, strategies, and processes. 

%% file: background.tex
\section{Background and Characteristics of Urban Transfer Learning}

\textit{Urban transfer learning}, as the technique aiming to address cold-start problems in smart city, is the intersection of two research areas: \textit{urban computing} \cite{zheng2014urban} and \textit{transfer learning} \cite{pan2010survey}.

\subsection{Urban Computing}

In reality, most urban computing applications can be categorized into the following aspects.

\textbf{Prediction:} Prediction in urban computing involves rich applications such as traffic demand \cite{zhang2017deep} and air quality \cite{zheng2013u} prediction. Generally, it involves two major types: (i) \textit{Fine-grained/Missing-value Prediction}: in urban monitoring tasks where the obtained data may not cover the whole city, we need to predict fine-grained data distribution based on the sparsely collected data \cite{zheng2013u,wang2016sparse}; (ii) \textit{Future Prediction}: with already collected data, we often need to predict the future data \cite{wei2016transfer,zhang2017deep}.

\textbf{Detection:} Detecting abnormal events or objects of interests is important in smart city services. For example, under destructive weather conditions such as typhoons and hurricanes, it is a critical issue to real-time detect road obstacles, such as fallen trees and ponding water; then, city authorities can restore road transportation in a timely manner to reduce losses \cite{chen2018radar}.

\textbf{Deployment:} Finding appropriate sites for deploying a new facility (e.g., shopping mall, electronic car charging station) is another major research topic \cite{cuo2018citytransfer}. It is worth noting that, once a facility is built, it will be difficult to move to other sites. In other words, the decision  of facility deployment cannot be undone. Hence, the mechanisms assisting facility deployment cannot adopt a trial-and-error methodology for iterative refinement, which makes it rather challenging.

\subsection{Transfer Learning}

In transfer learning, two key concepts deserve to be highlighted, i.e., \textit{source domain} and \textit{target domain}.

\textbf{Domain}. Briefly, \textit{domain} is a high-level concept that incorporates two components, i.e., a feature space $\mathcal X$ and its marginal probability distribution $P(\mathcal X)$. A traditional supervised learning task is conducted in one domain to infer some variable $y$ based on certain $x \in \mathcal X$, with training samples $\mathcal D = \{(x_1, y_1), \cdots, (x_n, y_n)\}$. Here, $y$ can have different definitions according to specific tasks (regression, classification, ranking, etc.). 
The objective of a task is to learn a function $f$ which is able to map $x$ to $y$ as accurately as possible.

\textbf{Source and Target Domains}. In reality, it is not always the case that adequate training samples exist. 
Transfer learning is introduced to address this problem by learning the function in the domain where training samples are few or even zero, called the \textit{target domain} $\mathcal D_t$, from another related domain where training samples are adequate, called the \textit{source domain} $\mathcal D_s$. The difficulty is that the source domain is usually different from the target domain from various aspects, such as feature sets, feature distributions, or tasks. 

\color{blue}
\textbf{Transfer Learning Method Types}. Various types of transfer learning approaches have been developed, such as \textit{instance-}, \textit{feature-}, and \textit{model-based} transfer \cite{pan2010survey}. Briefly, \textit{instance-based} transfer moves a subset of labeled instances from the source domain to the target domain; \textit{feature-based} transfer learns some feature representation from the source domain that is deemed to be beneficial for the target domain; \textit{model-based} transfer trains a machine learning model in the source domain, and then partially transfers the model (e.g., some parameters in the model) to the target domain.
\color{black}

\begin{figure*}[t]
	\begin{center}
		\includegraphics[width=.95\linewidth] {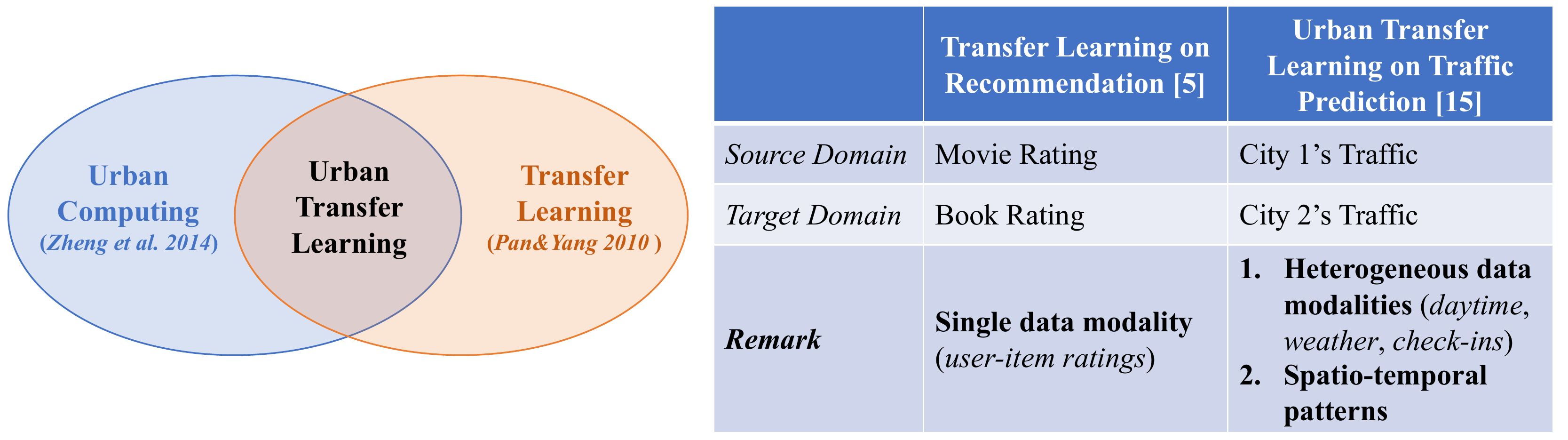}
		\caption{Relationship of \textit{urban transfer learning} and \textit{urban computing/transfer learning}. }
		\label{fig:urban_transfer}
	\end{center}
	\vspace{-1.5em}
\end{figure*}

\subsection{Characteristics of Urban Transfer Learning}
To date, much research effort has been devoted to transfer learning in the applications such as natural language processing and product recommendation \cite{pan2011domain,li2009can}, \color{blue} while urban transfer learning is still less studied. However, studying urban transfer learning is in fact not less important than other areas at all. For example, in recommendation systems, even if we adopt a simple strategy (e.g., recommending the most popular item) for cold-start new users without transfer learning, it usually has minor hurt since the recommendation performance will be soon improved as new users continue using the service. Comparatively, in urban applications such as chain store site selection in a new city, if a wrong site is determined and the store is built up, we cannot regret. Hence, we need to make a careful decision due to the high cost of a wrong decision. Then, transfer learning can play a crucial role in this decision making process~\cite{cuo2018citytransfer}. Particularly, directly applying existing transfer learning methods in smart city applications may not obtain the desirable results, as urban transfer learning has its distinct characteristics:

\color{black}
\textbf{Heterogeneous Data Modalities.} Traditional transfer learning often transfers knowledge between domains of the same data modality (e.g., between user ratings of movies and books). However, smart city applications are usually built on heterogeneous data with diverse formats. For example, air quality prediction is based on historical air quality records, digital map, points-of-interests, vehicle trajectories, etc.~\cite{zheng2013u}. To this end, cross-modality data fusion is a necessity in urban transfer learning but has not been well studied in transfer learning literature.

\textbf{Spatio-Temporal Patterns.} Smart city applications heavily depend on spatio-temporal datasets, such as vehicle trajectories, meteorology records, and urban events. In fact, spatio-temporal datasets often share a variety of patterns found by geography and statistics researchers, such as the \textit{Tobler's first law of geography} and the \textit{temporal trend, seasonal, and cyclic behaviors} \cite{hamilton1994time}. How to effectively leverage such spatio-temporal patterns  is still not well understood in transfer learning.

Figure~\ref{fig:urban_transfer} highlights the relationship between urban transfer learning and the two related research areas, and elaborates an example of an urban transfer learning application on traffic prediction, compared to a traditional transfer learning application on product recommendation.

%% file: what_to_transfer.tex
\section{What to Transfer in Urban Transfer Learning}

A key issue for transfer learning is `\textit{what to transfer?}' \cite{pan2010survey}. \color{blue} As a general rule, the two domains for knowledge transfer should be related, although not exactly the same. In practice, it is often empirically decided according to the application. 
For instance, in recommendation systems, knowledge can be transferred between different item categories (e.g., movie and book); in image recognition, recent studies usually transfer knowledge from the large-scale labeled image dataset like ImageNet (http://www.image-net.org/).
For urban transfer learning, considering the characteristics of `heterogeneous data modalities' and `spatio-temporal patterns', we categorize two useful transfer strategies, \textit{cross-modality} and \textit{cross-city}.

\color{black}
\textbf{Cross-Modality}. To cold-start a new smart city service, we can use data \textit{modalities} collected from existing services to learn the patterns in the new data \textit{modality} of the targeted service. For example, suppose we want to develop a new ridesharing service platform, but we do not have any data about the behaviors of ridesharing cars. Intuitively, ridesharing car behaviors could be similar to taxis. Then, to implement the services related to ridesharing, such as demand-supply prediction, we may leverage the existing data modalities collected from taxi services (e.g., taxi orders and trajectories). Nowadays, many city governments are publishing a large amount of data, such as NYC OpenData. This offers better cross-modality transfer opportunities to cold-start a new urban service. Besides, public online services may also generate beneficial source data modalities. One representative is the social network services such as Facebook and Twitter, where users' social posts and activities can be seen as useful proxies of urban dynamics. For instance, the popularity of social network check-ins may be an indicator of the density of physical crowd flows \cite{yang2016participatory}. Then, when we do not have adequate real crowd flow data, check-ins could be a proxy modality to realize transfer learning.

\textbf{Cross-City}. 
Another non-ignorable knowledge source for building a new smart city application is the experience from other cities with the same application already deployed, called \textit{source cities}. Generally, whether the cross-city transfer can work depends on the \textit{transferability of spatio-temporal patterns} of the target application. For example, the crowd flow dynamic patterns learned from the CBD (central business district) of a source city can probably benefit the crowd flow prediction service on the CBD of a target city, since human mobility  is highly related to city region functions~\cite{castro2013taxi}.  While the basic idea of cross-city transfer is intuitive, we emphasize that it can also face various difficulties in practice. The source and target cities may be quite different in population, development levels, etc., and even come from two distinct countries and cultures. This requires developing sophisticated transfer learning methods to avoid potential `\textit{negative transfer}'~\cite{pan2010survey} between cities. 

\textbf{Combining Cross-Modality and Cross-City Transfer}. In practice, cross-modality and cross-city transfer can be leveraged together. For instance, on one hand, we may learn the inter-modality correlations in one city, find the `invariant' knowledge in such correlations and transfer it to another city. Then, with the knowledge of inter-modality correlations, we can build inter-city relationships even some modality is missed. As a concrete example, suppose we want to build inter-city similarity on crowd flow dynamics but we cannot find enough crowd flow historical records for some cities, then we may rely on social media check-ins to construct the inter-city similarity, while the inter-modality correlation of check-in and crowd flow is actually learned from the cities with both rich historical check-in and crowd flow records. In such a way, we can transfer the crowd flow prediction model from one source city to a target new city (cross-city) with the help of check-ins as the proxy (cross-modality) \cite{wang2018crowd}. Later in this article, we will illustrate several urban transfer learning applications, and readers will see that most adopt both cross-modality and cross-city transfer.

%% file: applications.tex
\section{General Process of Urban Transfer Learning and Its Applications}

In this section, we demonstrate a general process framework for urban transfer learning. We then illustrate our three ongoing projects, so as to elaborate on how the framework can help design urban transfer learning applications in prediction, detection, and deployment, respectively. We expect that the general process framework can provide researchers with a systematic view of urban transfer learning applications.

\subsection{General Process Framework}

\begin{figure*}[t]
	\begin{center}
		\includegraphics[width=.95\linewidth] {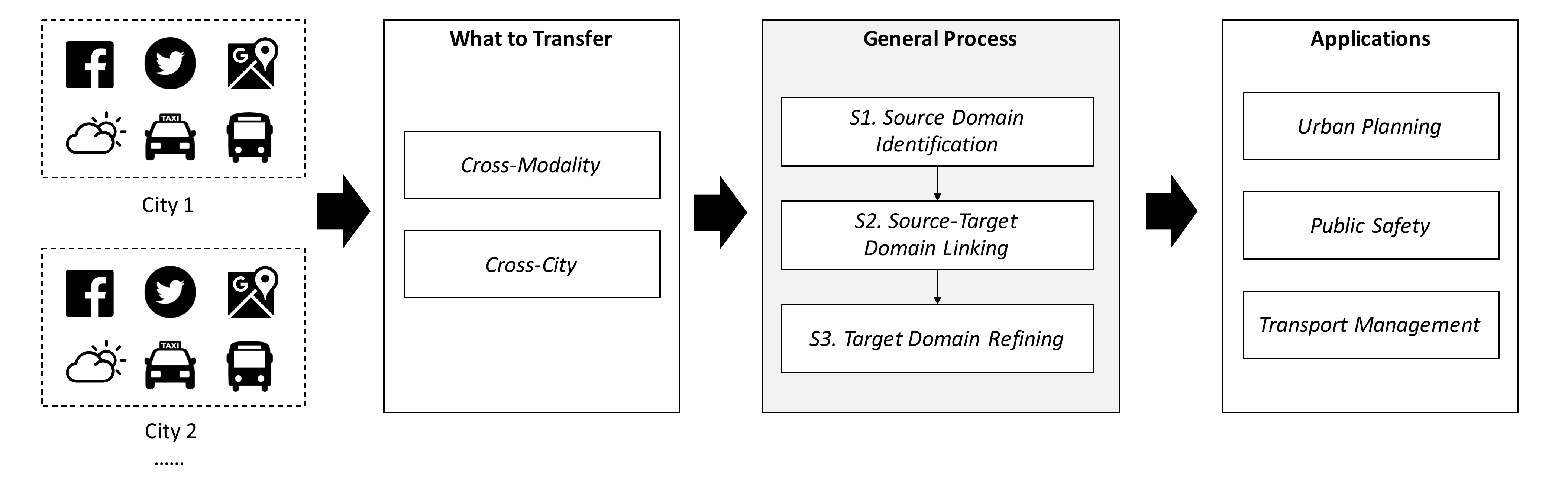}
		\vspace{-1em}
		\caption{Framework of urban transfer learning.}
		\label{fig:framework}
	\end{center}
	\vspace{-1.5em}
\end{figure*}

Figure~\ref{fig:framework} illustrates a general process framework for urban transfer learning including three steps: `\textit{S1. source domain identification}', `\textit{S2. source-target domain linking}', and `\textit{S3. target domain refining}'. Briefly, \textit{S1} determines the source domain and which part of knowledge should be transferred; \textit{S2} extracts the `invariant' part of the knowledge from the source domain and injects it to the target domain; \textit{S3} finally refines the transferred knowledge for the target application.

\textbf{S1. Source Domain Identification}. \textit{S1} determines the source domain and which part of knowledge can be transferred. While the previous section elaborates common ways to obtain  source domain knowledge, finding the most appropriate source domain still requires creativity and expertise. In practice, it is non-trivial to find a source domain to include all the desired information. To this end, we need to keep in mind that all the heterogeneous modalities of urban data in the target city, as well as the data from other cities, should be comprehensively considered as candidate parts of the source domain.

\textbf{S2. Source-Target Domain Linking}. \textit{S2} aims to extract the `invariant' part of the knowledge to bridge the source domain and the target domain. As aforementioned, instance-, feature-, and model-based methods may be designed here for effective knowledge transfer. Rather than only leveraging one type of the approaches, to achieve good performance in real applications, we may design a mechanism to integrate multiple types of transfer learning approaches.


\textbf{S3. Target Domain Refining}. While \textit{S2} has obtained useful knowledge from the source domain, directly and solely using it is usually not enough. Therefore, in addition to the knowledge transferred from the source domain, \textit{S3} tries to find more target-domain-specific characteristics and then incorporates such characteristics into the final model for the target application. According to different urban transfer learning scenarios, this refining process is varied. Generally, if we have a small number of labeled data in the target domain, then \textit{S3} will refine the final learned model to better fit the target labeled data. If no labeled data exists, more sophisticated mechanisms are needed: some of our attempts will be illustrated in the next a few subsections.

With this general process framework, we next elaborate three urban transfer learning applications in prediction, detection, and deployment, respectively.

\subsection{Application 1: Crowd Flow Prediction for Early Warning}

\begin{figure*}[t]
	\begin{center}
		\includegraphics[width=1\linewidth] {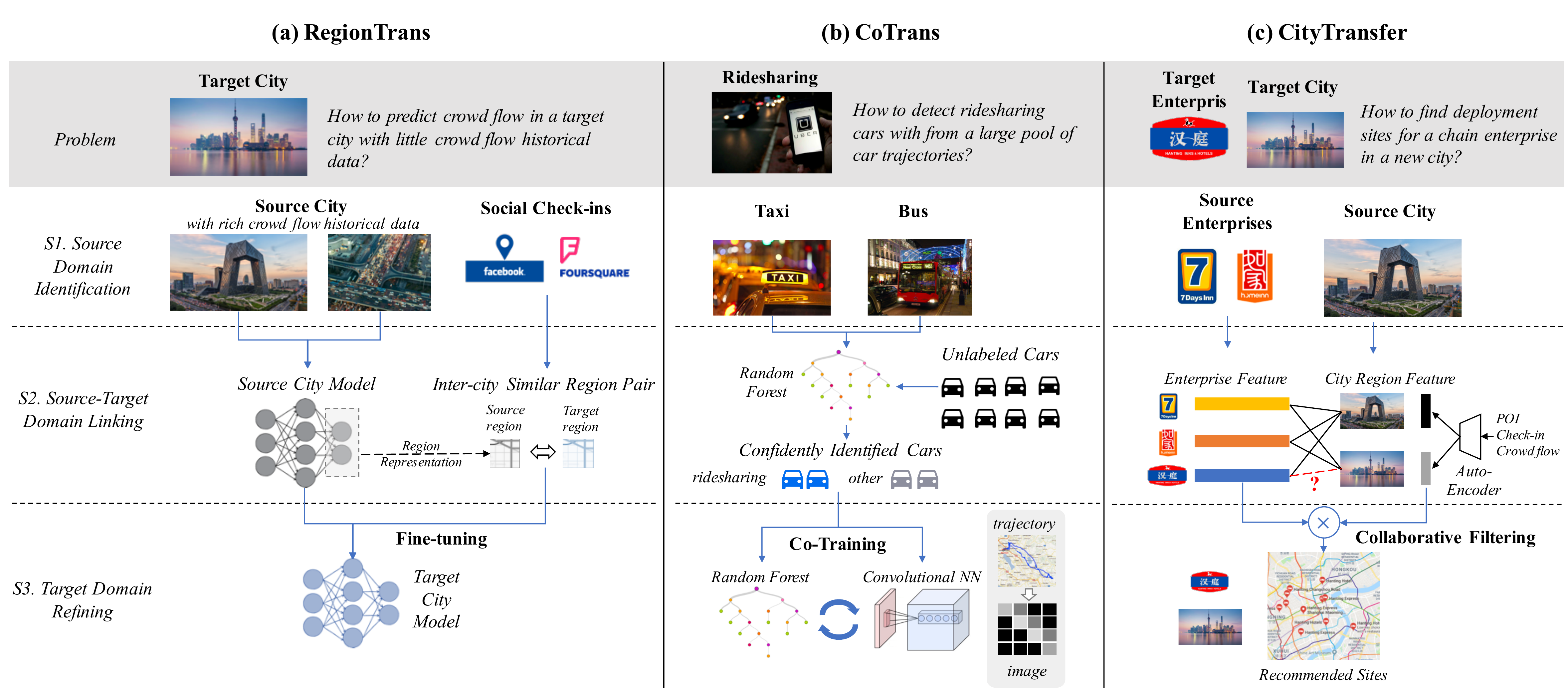}
		\vspace{-1em}
		\caption{Urban transfer learning applications. (a) \textit{RegionTrans}: crowd flow prediction in a new city; (b) \textit{CoTrans}: ridesharing detection without labeled data; (c) \textit{CityTransfer}: deployment site recommendation for chain enterprise in a new city.}
		\label{fig:app_all_in_one}
	\end{center}
	\vspace{-1.5em}
\end{figure*}


On 31 Dec. 2014,  around 300,000 people gathered near Chen Yi Square on the Bund, Shanghai for the New Year celebration, resulting in a deadly stampede that killed 36 people. To forecast such public safety risks, crowd flow prediction has attracted research efforts in recent years. Existing solutions usually assume that a city has a rich set of historical crowd flow records to train a prediction model \cite{zhang2017deep}, but this is not always the case. In this project, we aim to develop a cold-start crowd flow prediction solution for the cities which only hold a limited amount of historical crowd flow data. As deep leaning has become the state-of-the-art solution for crowd flow prediction \cite{zhang2017deep}, our focus is a deep transfer learning mechanism.

Figure \ref{fig:app_all_in_one} (a) shows the overall design of our mechanism \textit{RegionTrans} \cite{wang2018crowd}. Briefly, the applicability of \textit{RegionTrans} lies in the fact that there are usually similar regions between cities (e.g., CBD), and thus the crowd flow patterns of such inter-city similar regions can be transferred. More specifically, in \textit{S1}, we employ both \textit{cross-modality} and \textit{cross-city} strategies: we find \textit{a source city} with rich historical crowd flow data (e.g., several months) as one part of the source domain; in addition, we use \textit{social media check-ins} as a cross-modality proxy to measure the crowd flow similarity between city regions. In \textit{S2}, we design a deep spatio-temporal model which can extract region-level representation for crowd flow prediction. Then, for each region in the target city, we link it to the most similar regions in the source city so as to enable \textit{feature-based} transfer: the representations of inter-city similar regions are optimized to be similar. We also apply \textit{model-based} transfer: the neural network parameters learned on the source city are used as the parameter initialization for the target city. In \textit{S3}, with the limited amount of crowd flow data in the target city (e.g., one-day), we fine-tune the parameters of and obtain the final target city crowd flow prediction model. 

We evaluate \textit{RegionTrans} with the case of \textit{bike-sharing travel flow} \cite{zhang2017deep}. We choose two cities, Chicago and Washington D.C. as our experimental cities (one is the source and one is the target). Social media check-ins are collected from Foursquare \cite{yang2016participatory}. The results show that, compared to the state-of-the-art methods only considering the target city data, \textit{RegionTrans} can reduce the prediction error by up to 26\%~\cite{wang2018crowd}.

\subsection{Application 2: Ridesharing Detection for Unlicensed Car Regulation}
Ridesharing has become one of the major alternatives to travel in many cities, but it also incurs a black market that occasionally leads customers to risks. In May 2016, the driver of an unauthorized ridesharing car with a fake number plate robbed and killed a passenger in China. If we can detect cars suspected to be ridesharing but not licensed on the ridesharing platform, then city governors can take regulation actions more easily in time. Hence, this project aims to find ridesharing cars from a large number of candidate cars based on their trajectories \cite{wang2017ridesourcing}. The difficulty of ridesharing detection is the lack of historical trajectory data of ridesharing cars (i.e., lack of labeled data) because: (1) some cities do not officially allow ridesharing services yet; (2) even in the cities allowing ridesharing,  the trajectory data is held by companies (e.g., DiDi and Uber), which is not always accessible to governors. 


To address this difficulty, we develop a ridesharing detection mechanism called \textit{CoTrans} \cite{wang2017ridesourcing}, which can detect whether a car is ridesharing by transferring knowledge from taxis (Figure~\ref{fig:app_all_in_one} (b)). Taxis share many similar characteristics with ridesharing cars (e.g., driving distance and time) \cite{rayle2016just}, which makes this transfer feasible. In \textit{S1}, we use \textit{cross-modality} strategy by learning taxi patterns. Note that besides taxi trajectory data, we also include negative cases of non-taxi trajectory data such as buses and slag trucks, which are often accessible to city government offices such as \textit{Transportation Management} and \textit{Environmental Protection Agencies}. 
Based on these data, we train a classifier to identify taxis by driving distance, time, coverage, etc. with Random Forest (RF). In \textit{S2}, we use RF to classify cars in the candidate pool. This is a \textit{model-based} transfer as the classifier model RF learned on the source domain (taxi/non-taxi) is applied to the target domain (ridesharing/non-ridesharing). But directly using RF is not enough as taxis are not exactly the same as ridesharing. Hence, a more sophisticated mechanism is proposed to obtain more ridesharing-specific features. Particularly, we only keep the \textit{high-confidently} identified taxi/non-taxi in \textit{S2} (e.g., classification probability is $>0.9$), and label these cars as ridesharing/non-ridesharing. With such `\textit{pseudo-labeled}' ridesharing/non-ridesharing cars, \textit{S3} incorporates a co-training mechanism \cite{blum1998combining} where a new Convolutional Neural Network (CNN) classifier is built in addition to RF. The input of CNN is an image converted from the car's trajectory: if a car stages in a region for a longer time, the corresponding pixel is brighter (rightmost part of \textit{S3} in Figure~\ref{fig:app_all_in_one} (b)). In the co-training process, CNN and RF are refined collectively: the high-confidently detected ridesharing/non-ridesharing cars by CNN or RF are iteratively added to the `pseudo-labeled' instances to re-train both CNN and RF until convergence. Finally, the ensemble of CNN and RF is leveraged for ridesharing detection.

We evaluate \textit{CoTrans} on about 10,000 cars in Shanghai. 
The result shows that \textit{CoTrans} can achieve up to 85\% detection accuracy without the need of any labeled data, which is competitive to the accuracy of manual labels \cite{wang2017ridesourcing}. Hence, \textit{CoTrans} can serve as an automatic suspicious ridesharing car detection mechanism for city governors without the need for labeled ridesharing car data.

\subsection{Application 3: Deployment Recommendation for Chain Enterprise Extension}

In the final application, we illustrate the case that a chain enterprise (i.e., \textit{target enterprise}) wants to select appropriate sites for building its stores in a new city (i.e., \textit{target city}) to extend its business. As no existing chain store of the target enterprise has been built in the target city, this is a cold-starting smart city problem. Such a problem can often occur for chain businesses such as hotels and shopping malls.


To address this problem, we propose \textit{CityTransfer} \cite{cuo2018citytransfer}, as shown in Figure~\ref{fig:app_all_in_one}~(c). 
Briefly, \textit{CityTransfer} tackles the deployment problem with the collaborative filtering (CF) technique: we see a \textit{chain enterprise} as a \textit{user} $u$ and a \textit{city region} as an \textit{item} $i$. Then, \textit{CityTransfer} estimates a deployment score for any pair $\langle u_t, i_t \rangle$ where $u_t$ is the target enterprise and $i_t \in I_t$ is the set of regions in the target city. As no training data exists in the target city for the target enterprise, \textit{S1} first identifies a source domain with both \textit{cross-modality} and \textit{cross-city} knowledge. For \textit{cross-modality} transfer, we find other \textit{source chain enterprises} which already opened the business in the target city; for \textit{cross-city} transfer, we find the \textit{source city} where both source and target enterprises  have the business. The principal idea of transfer is to use CF to decompose and learn features for (source/target) enterprises $\mathbf u$ and (source/target) city regions $\mathbf i$, and then transfer the enterprise features $\mathbf u$ across cities. Note that when extracting region features $\mathbf i$ from data sources such as POIs and check-ins, the challenge is that different cities have diverse data distribution. Hence, in \textit{S2}, to embed the regions from different cities into a shared representation space, we propose an inter-city co-optimized AutoEncoder to generate a new representation space based on raw features of POIs, check-ins, etc., for \textit{feature-based} transfer. The new space is optimized to map similar inter-city region pairs to similar features. 
Finally, in \textit{S3}, with the shared feature space, we can learn the target enterprise feature $\mathbf u_t$ and the target city region feature $\mathbf i_t$, which can further infer the deployment scores for the target enterprise in the target city.

We conduct experiments on three popular chain hotel enterprises in China, i.e., \textit{Hanting}, \textit{7 Days}, and \textit{Home Inn}. With Beijing as the source city, we run tests by transferring the knowledge to Shanghai, Xi'an, and Nanjing for each company while seeing the other companies as sources. Results show that the hotel deployment sites selected by \textit{CityTransfer} can attract more customers than the sites recommended by traditional empirical methods with only local city features such as crowd flow \cite{cuo2018citytransfer}.

\color{blue}
\subsection{Summary}

\begin{figure*}[t]
	\begin{center}
		\includegraphics[width=.9\linewidth] {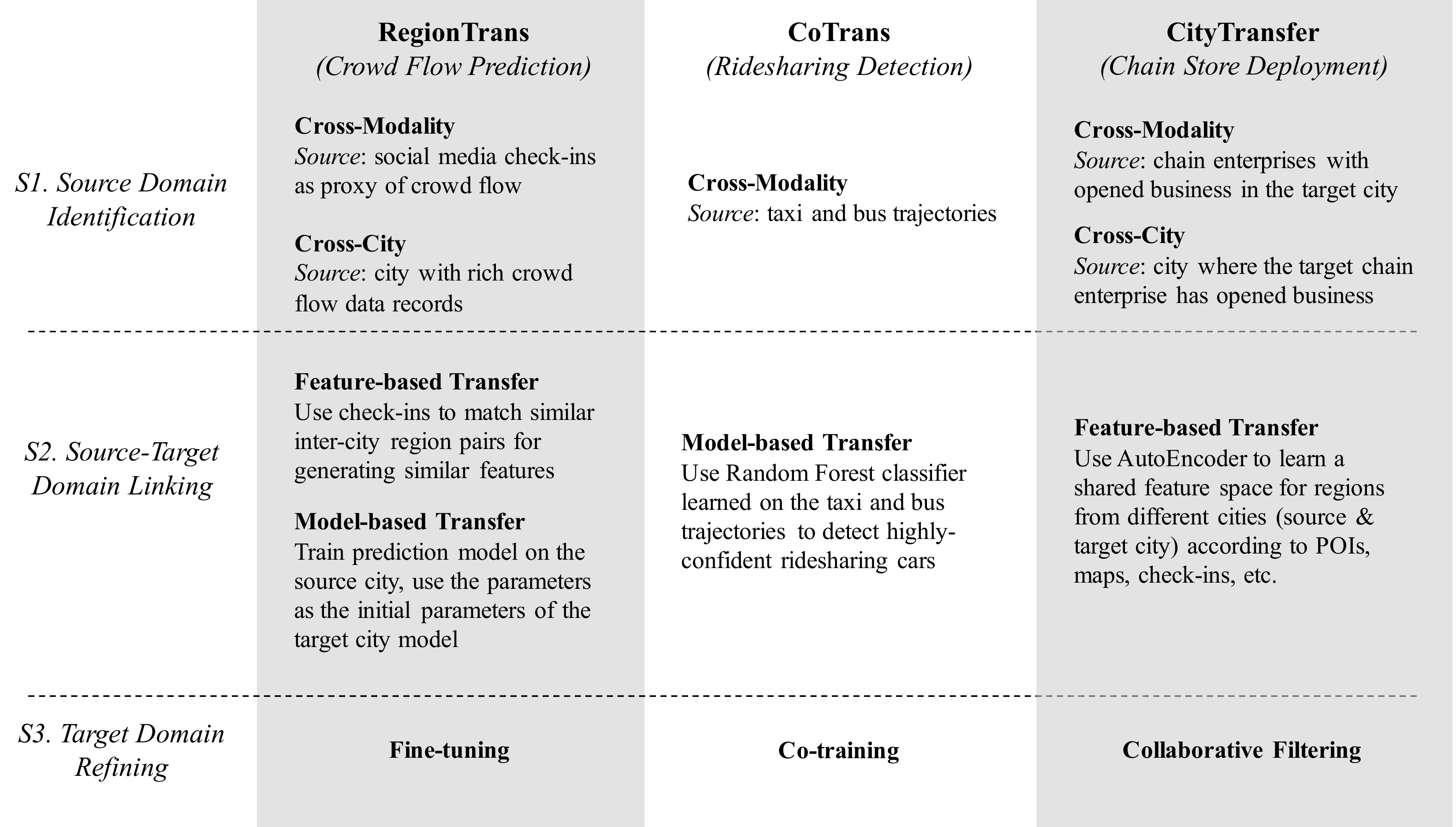}
		\caption{Summary of three urban transfer learning applications.}
		\label{fig:summary_of_app}
	\end{center}
	\vspace{-1.5em}
\end{figure*}

As shown in Figure~\ref{fig:app_all_in_one}, three urban transfer learning applications can fit our proposed framework very well. We further summarize their characters in Figure~\ref{fig:summary_of_app}. It shows that in each step of our framework, different methods can be developed to achieve the objective of the corresponding step. Regardless of method details, \textit{cross-modality} and \textit{cross-city} are main knowledge transfer strategies. We expect that our illustrated three applications can serve as reference algorithms and inspire researchers to build their own urban transfer learning algorithms.

\color{black}

%% file: challenges.tex
\section{Future Research Opportunities}

Urban transfer learning is still at a preliminary research stage where a few opportunities exist. We expect that this article can attract more research efforts into this research area.

\color{blue}
\textbf{Adversarial Neural Network.} The adversarial neural network is a quickly developing technique and it has been successfully used in transfer learning \cite{Ganin2016DomainAdversarialTO}. This may be a potential direction for developing efficient urban transfer learning algorithms. For example, in the cross-city transfer, it is possible that adversarial network can be leveraged to extract the features that cannot discriminate between cities, leading to better transferability.
\color{black}

\textbf{Exploring More Source Knowledge.} Besides \textit{cross-modality} and \textit{cross-city}, there may be more transfer opportunities. For example, many urban phenomena are related to urban events (e.g., festivals). Suppose that a city plans to hold a big event (e.g., Olympic games) for the first time, no previous experience can directly help build service for the event. However, if the city previously held other big events (e.g., World Cup Football), then some kind of \textit{cross-event} transfer may be possible.

\textbf{Assessing Transferability and {\color{blue}Avoiding Negative Transfer}.}  A fundamental challenge is quantitatively measuring the transferability between the source and target domain. Take cross-city transfer as an example, suppose that we have several source city candidates, then assessing the transferability will help to select the appropriate cities as the final source cities. 
\color{blue} 
Another important usage of transferability assessment is to avoid negative transfer. Most of today's transfer learning applications rely on trial-and-error to judge whether the transfer is effective. However, some smart city applications like deployment cannot be learned by trial-and-error. Transferability assessment can then help decide `what to transfer' for such applications.

\color{black}
\textbf{Dealing with Privacy-Preserving Data}. Recently, more and more urban data is published in a privacy-preserving manner to avoid potential privacy breaches to citizens. For example, the taxi trip records published by many cities only include coarse pick-up and drop-off regions rather than GPS coordinates. How to efficiently leverage privacy-preserving data becomes a new challenge.
\color{blue}

\textbf{Urban Multi-Task Learning}. Multi-task learning is a special type of transfer learning, where tasks in different domains are learned simultaneously~\cite{Zhang2017ASO}. Although this paper does not put a focus on multi-task learning, we believe that our framework and guidelines are still helpful. For example, from the perspectives of cross-modality and cross-city, we can probably find the multiple tasks that can be learned together. Besides, urban multi-task learning may enable a healthy data sharing environment, as each city does not only give out data but also obtains benefits from the other cities' data.

\vspace{+2em}
\textit{Acknowledgment}:  This research is partially supported by the National Natural Science Foundation of China (No. 61772045), the National Key R\&D Program of China (2017YFB1001800), and the National Natural Science Foundation of China (No. 61772428). The source code of \textit{RegionTrans} can be found in https://github.com/Di-Chai/RegionTransfer.


\color{black}